\documentclass{article}
\usepackage{ijcai17}

\usepackage{times}
\usepackage{epsfig}
\usepackage{graphicx}
\usepackage{amsmath}
\usepackage{amssymb}
\usepackage{array}
\usepackage{subfigure}
\usepackage{spverbatim}
\usepackage[ruled,vlined]{algorithm2e}
\usepackage{booktabs}
\usepackage{bm}
\usepackage{color}
\usepackage{float}
\usepackage{longtable}

\newcommand{\etal}{\emph{et al.}}

\title{IJCAI--17 Formatting Instructions}
\author{Zhuotun Zhu, Lingxi Xie, Alan Yuille\\
Johns Hopkins University, Baltimore, MD, USA  \\
\texttt{\{zhuotun, 198808xc, alan.l.yuille\}@gmail.com}}

\begin{document}

\title{Object Recognition with and without Objects}


\maketitle

\begin{abstract}
  While recent deep neural networks have achieved a promising performance on object recognition, they rely \textbf{\textit{implicitly}} on the visual contents of the whole image. In this paper, we train deep neural networks on the foreground (object) and background (context) regions of images respectively. Considering human recognition in the same situations, networks trained on the pure background \textbf{\textit{without}} objects achieves highly reasonable recognition performance that beats humans by a large margin if only given context. However, humans still outperform networks \textbf{\textit{with}} pure object available, which indicates networks and human beings have different mechanisms in understanding an image.  Furthermore, we straightforwardly combine multiple trained networks to explore different visual cues learned by different networks. Experiments show that useful visual hints can be \textbf{\textit{explicitly}} learned separately and then combined to achieve higher performance, which verifies the advantages of the proposed framework.
\end{abstract}

\section{Introduction}
\label{Introduction}
Object recognition is a long-lasting battle in computer vision,
which aims to categorize an image according to the visual contents.
In recent years, we have witnessed an evolution in this research field.
Thanks to the availability of large-scale image datasets~\cite{Deng_2009_ImageNet} and powerful computational resources,
it becomes possible to train a very deep convolutional neural network (CNN)~\cite{Krizhevsky_2012_ImageNet},
which is much more efficient beyond the conventional Bag-of-Visual-Words (BoVW) model~\cite{Csurka_2004_Visual}.

It is known that an image contains both foreground and background visual contents.
However, most object recognition algorithms focus on recognizing the visual patterns only on the foreground region~\cite{Zeiler_2014_Visualizing}.
Although it has been proven that background (context) information also helps recognition~\cite{Simonyan_2015_Very},
it still remains unclear if a deep network can be trained individually to learn visual information only from the background region.
In addition, we are interested in exploring different visual patterns by training neural networks on foreground and background separately for object recognition, which is less studied before.

\begin{figure}
  \centering
  \includegraphics[width=0.5\textwidth]{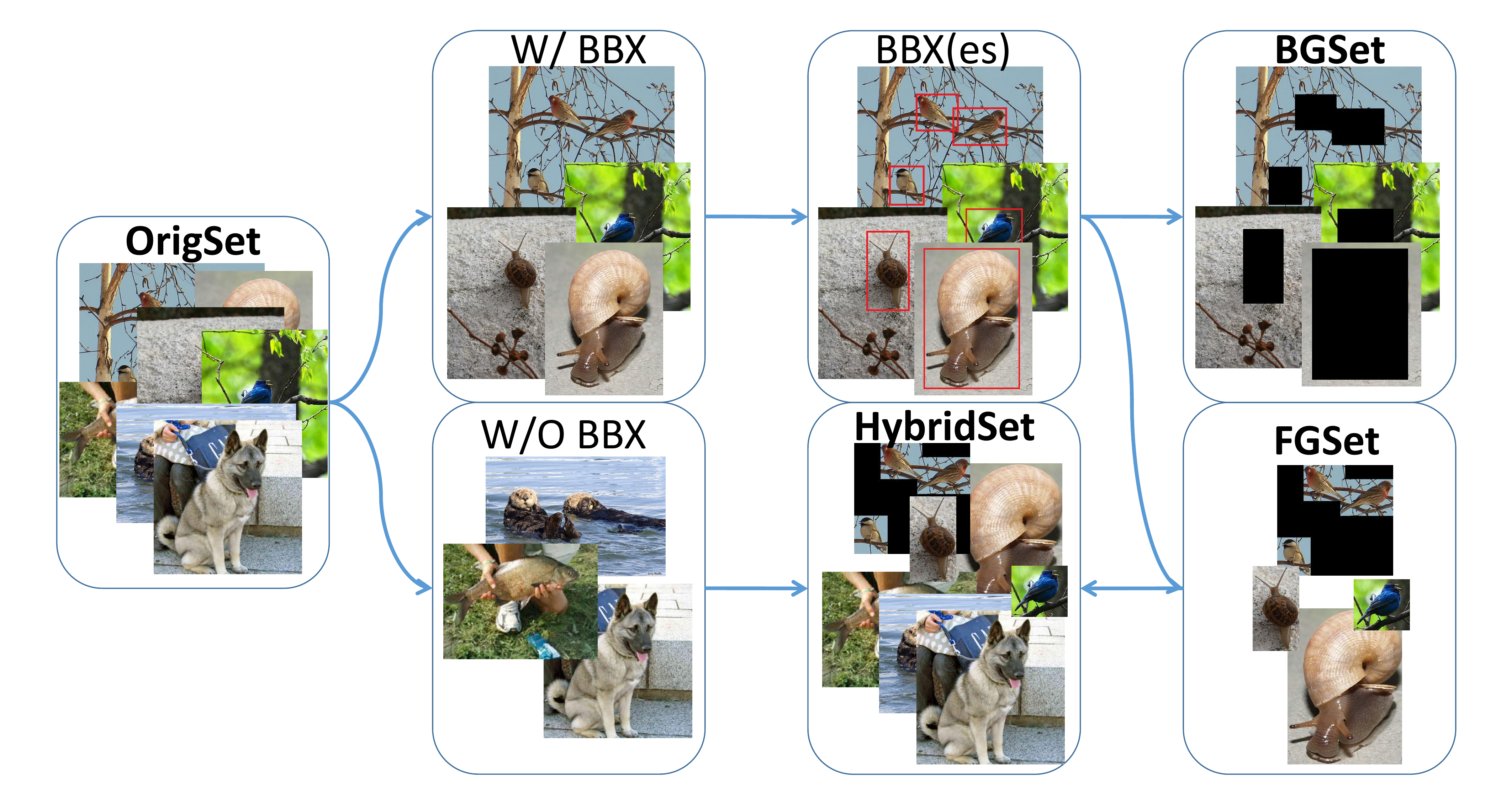}\\
  \caption{Procedures of dataset generation. First, we denote the original set as the {\bf OrigSet}, divided into two sets, one with the ground-truth bounding box (W/ BBX) and the other one without (W/O BBX). Then the set with labelled bounding box(es) are further processed by setting regions inside all ground-truth to be $0$'s to compose the {\bf BGSet} while cropping the regions out to produce the {\bf FGSet}. In the end, add the images without bounding boxes with {\bf FGSet} to construct the {\bf HybridSet}. Please note that some images of the {\bf FGSet} have regions to be black ($0$'s) since these images are labelled with multiple objects belonging to the same class, which are cropped according to the smallest rectangle frame that includes all object bounding boxes in order to keep as less background information as possible on {\bf FGSet}. Best viewed in color.}\label{Fig:Dataset}
\end{figure}

In this work, we investigate the above problems by explicitly training multiple networks for object recognition.
We first construct datasets from {\bf ILSVRC2012}~\cite{Russakovsky_2015_ImageNet}, {\em i.e.}, one {\em foreground} set and one {\em background} set, by taking advantage of the ground-truth bounding box(es) provided in both training and testing cases.
After dataset construction, we train deep networks individually to learn foreground (object) and background (context) information, respectively.
We find that, even \textbf{\textit{only}} trained on pure background contexts, the deep network can still converge and makes reasonable prediction ($\bm{14.4}\%$ top-$1$ and nearly $\bm{30}\%$ top-$5$ classification accuracy on the background validation set).
To make a comparison, we are further interested in the human recognition performance on the constructed datasets.
Deep neural networks outperform non-expert humans in fine-grained recognition,
and humans sometimes make errors because they cannot memorize all categories of datasets~\cite{Russakovsky_2015_ImageNet}.
In this case, to more reasonably compare the recognition ability of humans and deep networks,
we follow~\cite{Huh_2016_What} to merge all the $1\rm{,}000$ fine-grained categories of the original {\bf ILSVRC2012},
resulting in a $127$-class recognition problem meanwhile keeping the number of training/testing images unchanged.
We find that human beings tend to pay more attention to the object while networks put more emphasis on context than humans for classification.
By visualizing the patterns captured by the background net, we find that some visual patterns are not available in the foreground net.
Therefore, we apply networks on the foreground and background regions respectively via the given ground-truth bounding box(es) or extracting object proposals without available ones.
We find that the linear combination of multiple neural networks can give higher performance.

To summarize, our main contributions are three folds:
1) We demonstrate that learning foreground and background visual contents \textit{\textbf{separately}} is beneficial for object recognition. 
Training a network based on pure background although being wired and challenging, is technically feasible and captures highly useful visual information.
2) We conduct \textit{\textbf{human recognition}} experiments on either pure background or foreground regions to find that human beings outperform networks on pure foreground while are beaten by networks on pure background, which implies the different mechanisms of understanding an image between networks and humans.
3) We straightforwardly \textit{\textbf{combine}} multiple neural networks to explore the effectiveness of different learned visual clues under two conditions \textit{\textbf{with}} and \textit{\textbf{without}} ground-truth bounding box(es), which gives promising improvement over the baseline deep neural networks.

\section{Related Work}
\label{RelatedWork}

Object recognition is fundamental in computer vision field, which is aimed to understand the semantic meaning among an image via analyzing its visual contents.
Recently, researchers have extended the traditional cases~\cite{Lazebnik_2006_Beyond}
to fine-grained~\cite{Wah_2011_Caltech}~\cite{Nilsback_2008_Automated}~\cite{lin2015bilinear}, and large-scale~\cite{Xiao_2010_SUN}~\cite{Griffin_2007_Caltech} tasks.
Before the exploding development of deep learning, the dominant BoVW model~\cite{Csurka_2004_Visual} represents every single image with a high-dimensional vector.
It is typically composed of three consecutive steps, {\em i.e.}, descriptor extraction~\cite{Lowe_2004_Distinctive}~\cite{Dalal_2005_Histograms}, feature encoding~\cite{Wang_2010_Locality}~\cite{Perronnin_2010_Improving} and feature summarization~\cite{Lazebnik_2006_Beyond}.



The milestone Convolutional Neural Network (CNN) is treated as a hierarchical model for large-scale visual recognition.
In past years, neural networks have already been proved to be effective for simple recognition tasks~\cite{LeCun_1990_Handwritten}.
More recently, the availability of large-scale training data ({\em e.g.}, ImageNet~\cite{Deng_2009_ImageNet}) and powerful computation source like GPUs make it practical to train deep neural networks~\cite{Krizhevsky_2012_ImageNet}~\cite{zhu2016deep} which significantly outperform the conventional models.
Even deep features have been proved to be very successful on vision tasks like object discovery~\cite{wang2015relaxed}, object recognition~\cite{xietowards}, etc.
A CNN is composed of numerous stacked layers, in which responses from the previous layer are then convoluted and activated by a differentiable function, followed by a non-linear transformation~\cite{Nair_2010_Rectified} to avoid over-fitting.
Recently, several efficient methods were proposed to help CNNs converge faster and prevent over-fitting~\cite{Krizhevsky_2012_ImageNet}.
It is believed that deeper networks produce better recognition results~\cite{Szegedy_2015_Going}\cite{Simonyan_2015_Very},
but also requires engineering tricks to be trained very well~\cite{Ioffe_2015_Batch}~\cite{He_2016_Deep}.

Very few techniques on background modeling~\cite{bewley2017background} have been developed for object recognition, despite the huge success of deep learning methods on various vision tasks.~\cite{shelhamer2016fully} proposed the fully convolutional networks (FCN) for semantic segmentation, which are further trained on foreground and background defined by shape masks. They find it is not vital to learn a specifically designed background model. For face matching,~\cite{sanderson2009multi} developed methods only on the cropped out faces to alleviate the possible correlations between faces and their backgrounds.~\cite{han2015background} modeled the background in order to detect the salient objects from the background.~\cite{doersch2014context} showed using the object patch to predict its context as supervisory information can help discover object clusters, which is consistent with our motivation to utilize the pure context for visual recognition. To our best knowledge, we are the first to explicitly learn both the foreground and background models and then combine them together to be beneficial for the object recognition.


Recently, researchers pay more attention to human experiments on objects recognition. Zhou \etal~\cite{zhou2014object} invited Amazon Mechanical Turk (AMT) to identify the concept for segmented images with objects. They found that the CNN trained for scene classification automatically discovers meaningful object patches. While in our experiments, we are particularly interested in the different emphasis between human beings and networks for recognition task.


Last but not the least, visualization of CNN activations is an effective method to understand the mechanism of CNNs.
In~\cite{Zeiler_2014_Visualizing}, a {\em de-convolutional} operation was proposed
to capture visual patterns on different layers of a trained network.
\cite{Simonyan_2015_Very} and~\cite{Cao_2015_Look} show that different sets of neurons are activated
when a network is used for detecting different visual patterns.
In this work, we will use a much simpler way of visualization which is inspired by~\cite{Wang_2015_Discovering}.

\section{Training Networks}
\label{ComplementaryNets}

Our goal is to explore the possibility and effectiveness of training networks on foreground and background regions, respectively.
Here, foreground and background regions are defined by the annotated ground-truth bounding box(es) of each image.
All the experiments are done on the datasets composed from the {\bf ILSVRC2012}.

\begin{table*}
\small
\begin{center}
\begin{tabular}{lcrcc}\toprule
Dataset         & Image Description                  & \# Training Image  & \# Testing Image & Testing Accuracy\\
\hline
{\bf OrigSet}   & Original Image                     & $1\rm{,}281\rm{,}167$ & $50\rm{,}000$      & $58.19\%$, $80.96\%$ \\
{\bf FGSet}     & Foreground Image                   &        $544\rm{,}539$ & $50\rm{,}000$      & $60.82\%$, $83.43\%$ \\
{\bf BGSet}     & Background Image                   &        $289\rm{,}031$ & $50\rm{,}000$      & $14.41\%$, $29.62\%$ \\
{\bf HybridSet} & Original Image or Foreground Image & $1\rm{,}281\rm{,}167$ & $50\rm{,}000$   & $61.29\%$, $83.85\%$    \\
\bottomrule
\end{tabular}
\end{center}
\caption{
    The configuration of different image datasets originated from the {\bf ILSVRC2012}. The lass column denotes the testing performance of trained {\bf AlexNet} in terms of top-$1$ and top-$5$ classification accuracy on corresponding datasets, {\em e.g.}, the {\bf BGNet} gives $14.41\%$ top-$1$ and $29.62\%$ top-$5$ accuracy on the testing images of {\bf BGSet}.
}\label{Tab:Dataset}
\end{table*}

\subsection{Data Preparation}
\label{ComplementaryNets:DataPreparation}

The {\bf ILSVRC2012} dataset~\cite{Russakovsky_2015_ImageNet}
contains about $1.3\mathrm{M}$ training and $50\mathrm{K}$ validation images.
Throughout this paper, we refer to the original dataset as {\bf OrigSet} and the validation images are regarded as our testing set.
Among {\bf OrigSet}, $544\rm{,}539$ training images and all $50\rm{,}000$ testing images are labeled with at least one ground-truth bounding box.
For each image, there is only one type of object annotated according to its ground-truth class label.

We construct three variants of training sets and two variants of testing sets from {\bf OrigSet} by details below.
An illustrative example of data construction is shown in Fig.~\ref{Fig:Dataset}.
The configuration of different image datasets are summarized in Table~\ref{Tab:Dataset}.

\begin{itemize}
\item The foreground dataset ({\bf FGSet}) is composed of all images with at least one available ground-truth bounding box.
For each image, we first compute the smallest rectangle frame which includes all object bounding boxes, then based on which the image inside the frame is cropped to be used as the training/testing data.
Note that if an image has multiple object bounding boxes belonging to the same class, we set all the background regions inside the frame to be $0$'s to keep as little context as possible on {\bf FGSet}.
There are totally $544\rm{,}539$ training images and $50\rm{,}000$ testing images on {\bf FGSet}.
Since the annotation is on the bounding box level, images of the {\bf FGSet} may contain some background information.
\item The construction of the background dataset ({\bf BGSet}) consists of two stages.
First, for each image with at least one ground-truth bounding box available, regions inside every ground-truth bounding box are set to $0$'s.
Chances are that almost all the pixels of one image are set to 0s if its object consists of nearly 100 percent of its whole region.
Therefore during training, we discard those samples with less than $50\%$ background pixels preserved,
{\em i.e.}, the {\em foreground frame} is larger than half of the entire image,
so that we can maximally prevent using those less meaningful background contents (see Fig~\ref{Fig:Dataset}).
However in testing, we keep all the processed images, in the end, $289\rm{,}031$ training images and $50\rm{,}000$ testing images are preserved.
\item To increase the amount of training data for foreground classification,
we also construct a hybrid dataset, abbreviated as the {\bf HybridSet}.
The {\bf HybridSet} is composed of all images of the original training set.
If at least one ground-truth bounding box is available, we pre-process this image as described on {\bf FGSet},
otherwise, we simply keep this image without doing anything.
As bounding box annotation is available in each testing case, the {\bf HybridSet} and the {\bf FGSet} contain the same testing data.
Training with the {\bf HybridSet} can be understood as a semi-supervised learning process.
\end{itemize}

\subsection{Training and Testing}
\label{ComplementaryNets:TrainingTesting}
We trained the milestone {\bf AlexNet}~\cite{Krizhevsky_2012_ImageNet} using the CAFFE library~\cite{Jia_2014_CAFFE} on different training sets as mentioned in the Sec~\ref{ComplementaryNets:DataPreparation}.

The base learning rate is set to $0.01$, and reduced by $1/10$ for every $100\rm{,}000$ iterations.
The moment is set to be $0.9$ and the weight decay parameter is $0.0005$.
A total number of $450\rm{,}000$ iterations is conducted, which corresponds to around $90$ training epochs on the original dataset.
Note that both {\bf FGSet} and {\bf BGSet} contain less number of images than that of {\bf OrigSet} and {\bf HybridSet},
which leads to a larger number of training epochs, given the same training iterations.
In these cases, we adjust the dropout ratio as $0.7$ to avoid the overfitting issue.
We refer to the network trained on the {\bf OrigSet} as the {\bf OrigNet},
and similar abbreviated names also apply to other cases, {\em i.e.}, the {\bf FGNet}, {\bf BGNet} and {\bf HybridNet}.


During testing, we report the results by using the common data augmentation of averaging 10 patches from the {\em $5$ crops} and {\em $5$ flips}. After all forward passes are done, the average output on the final ({\em fc-8}) layer is used for prediction. We adopt the MatConvNet~\cite{Vedaldi_2015_Matconvnet} platform for performance evaluation.

\vspace{-0.1cm} 
\section{Experiments}
\label{ComplementaryNets:PreliminaryResults}
The testing accuracy of {\bf AlexNet} trained on corresponding dataset are given in the last column of Table~\ref{Tab:Dataset}.
We can find that the {\bf BGNet} produces reasonable classification results: $14.41\%$ top-$1$ and $29.62\%$ top-$5$ accuracy (while the random guess gets $0.1\%$ and $0.5\%$, respectively), which is a bit surprising considering it makes classification decisions only on background contents \textit{\textbf{without}} any foreground objects given.
This demonstrates that deep neural networks are capable of learning pure contexts to infer objects even being fully occluded. Not surprisingly, the {\bf HybridNet} gives better performance than the {\bf FGNet} due to more training data available.

\subsection{Human Recognition}
\label{HumanRecognition}
As stated before, to alleviate the possibility of wrongly classifying images for humans beings due to high volume of classes up to $1\rm{,}000$ on the original {\bf ILSVRC2012}, we follow~\cite{Huh_2016_What} by merging all the fine-grained categories, resulting in a $127$-class recognition problem meanwhile keeping the number of training/testing images unchanged.
To distinguish the merged $127$-class datasets with the previous datasets, we refer to them as the {\bf OrigSet-127},  {\bf FSet-127} and {\bf BGSet-127}, respectively.
Then we invite volunteers who are familiar with the merged $127$ classes to perform the recognition task on {\bf BGSet-127} and {\bf FSet-127}. Humans are given $256$ images covering all $127$ classes and one image takes around two minutes to make the top-$5$ decisions. We do not evaluate humans on {\bf OrigSet-127} since we believe humans can perform well on this set like on {\bf OrigSet}. Human performance on {\bf OrigSet} (labeled by $^\star$) is reported by~\cite{Russakovsky_2015_ImageNet}.

Table~\ref{Tab:ResultsILSVRC} gives the testing recognition performance of human beings and trained {\bf AlexNet} on different datasets. It is well noted that humans are good at recognizing natural images~\cite{Russakovsky_2015_ImageNet}, {\em e.g.}, on {\bf OrigSet}, human labelers achieve much higher performance than {\bf AlexNet}. We can find the human beings also surpass networks on the foreground (object-level) recognition by $5.93\%$ and $1.96\%$ in terms of top-$1$ and top-$5$ accuracy. Surprisingly, {\bf AlexNet} beats human labelers to a large margin on the background dataset {\bf BGSet-127} considering the $127\%$ and $85\%$ relative improvements from $18.36\%$ to $41.65\%$ and $39.84\%$ to $73.79\%$ for top-$1$ and top-$5$ accuracy, respectively. In this case, the networks are capable of exploring background hints for recognition much better than human beings.
On the contrary, humans classify images mainly based on the visual contents of the foreground objects.

\begin{table}
\footnotesize
\begin{center}
\begin{tabular}{lcc}\toprule
Dataset           & {\bf AlexNet}        & {\bf Human}                \\
\hline
{\bf OrigSet}  & $58.19\%$, $80.96\%$ & $-$,       $94.90\%^\star$ \\
{\bf BGSet}    & $14.41\%$, $29.62\%$ & $-$,       $-$             \\
\hline
{\bf OrigSet-127} & $73.16\%$, $93.28\%$ & $-$,       $-$             \\
{\bf FGSet-127} & $75.32\%$, $93.87\%$ & $81.25\%$,       $95.83\%$             \\
{\bf BGSet-127}   & $41.65\%$, $73.79\%$ & $18.36\%$, $39.84\%$       \\
\bottomrule
\end{tabular}
\end{center}
\caption{
    Classification accuracy (in terms of top-$1$, top-$5$) on five sets by deep neural networks and human, respectively.
}
\label{Tab:ResultsILSVRC}
\end{table}

\subsection{Cross Evaluation}
\label{ComplementaryNets:CrossEvaluation}

\begin{table}
\footnotesize
\begin{center}
\resizebox{1.05\columnwidth}{!}{
\begin{tabular}{p{1cm}<{\raggedright}p{2.4cm}<{\centering}p{2.4cm}
<{\centering}p{2.4cm}<{\centering}}\toprule
Network         & {\bf OrigSet}        & {\bf FGSet}          & {\bf BGSet}          \\
\hline
{\bf OrigNet}   & $\bm{58.19\%}$, $\bm{80.96\%}$ & $50.73\%$, $74.11\%$ & $ 3.83\%$, $ 9.11\%$ \\
{\bf FGNet}     & $33.42\%$, $53.72\%$ & $60.82\%$, $83.43\%$ & $ 1.44\%$, $ 4.53\%$ \\
{\bf BGNet}     & $ 4.26\%$, $10.73\%$ & $ 1.69\%$, $ 5.34\%$ & $\bm{14.41\%}$, $\bm{29.62\%}$ \\
{\bf HybridNet} & $52.89\%$, $76.61\%$ & $\bm{61.29\%}$, $\bm{83.85\%}$ & $ 3.48\%$, $ 9.05\%$ \\
\bottomrule
\end{tabular}
}
\end{center}
\caption{
    Cross evaluation accuracy (in terms of top-$1$, top-$5$) on four networks and three testing sets.
    Note that the testing set of {\bf HybridSet} is identical to that of {\bf FGSet}.
}
\label{Tab:CrossEvaluation}
\end{table}

To study the difference in visual patterns learned by different networks,
we perform the cross evaluation, {\em i.e.}, applying each trained network to different testing sets.
Results are summarized in Table~\ref{Tab:CrossEvaluation}.

We find that the transferring ability of each network is limited,
since a model cannot obtain satisfying performance in the scenario of different distributions between training and testing data.
For example, using {\bf FGNet} to predict {\bf OrigSet} leads to $27.40\%$ absolute drop ($45.05\%$ relative) in top-$1$ accuracy,
meanwhile using {\bf OrigNet} to predict {\bf FGSet} leads to $7.46\%$ drop ($12.82\%$ relative) in top-$1$ accuracy.
We conjecture that {\bf FGNet} may store very little information on contexts, thus confused by the background context of {\bf OrigSet}.
On the other side, {\bf OrigNet} has the ability of recognizing contexts but is wasted for the task on {\bf FGSet}.


\begin{figure}[ht]
\centering
\subfigure{
    \includegraphics[width=0.4\textwidth]{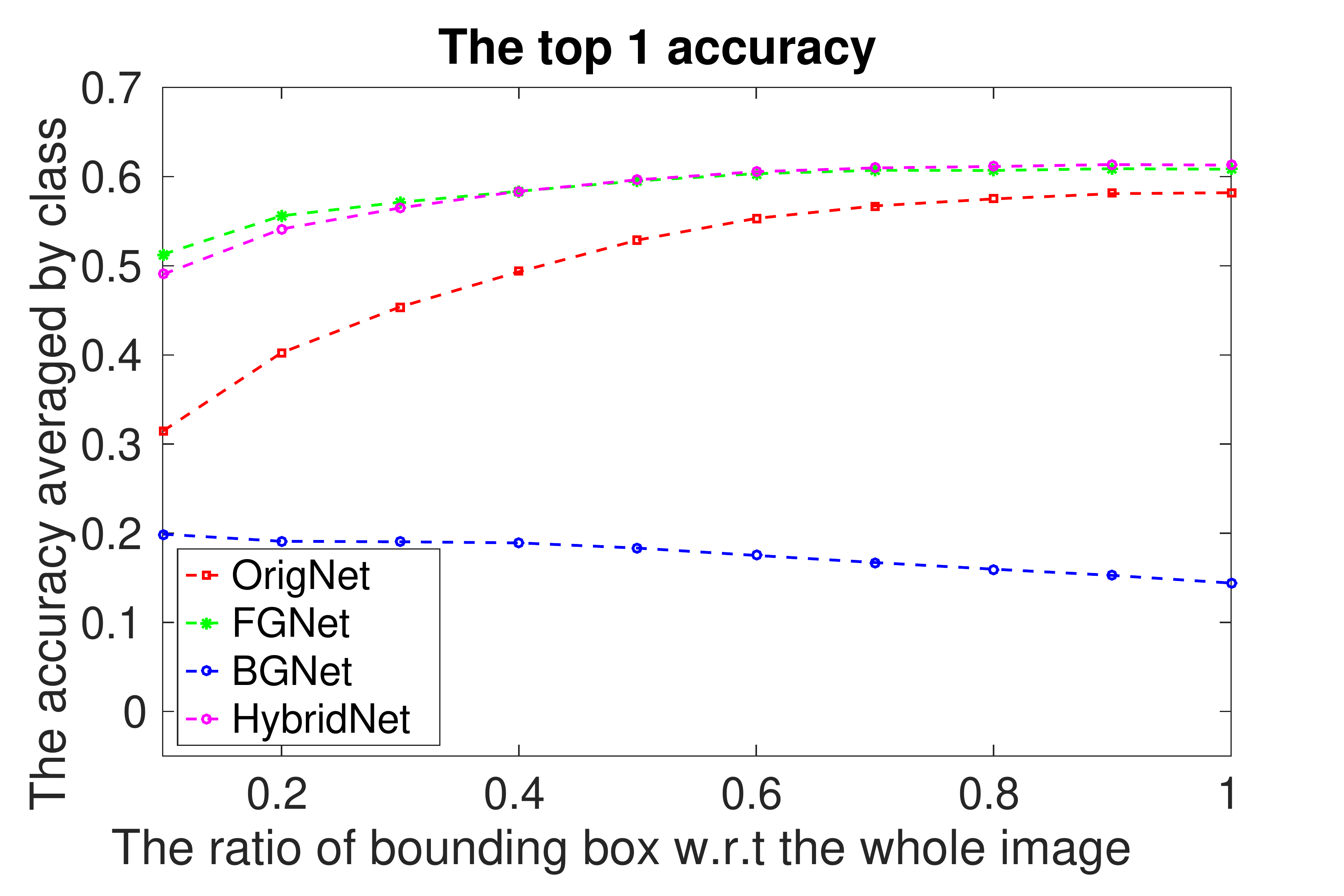}
}
\subfigure{
    \includegraphics[width=0.4\textwidth]{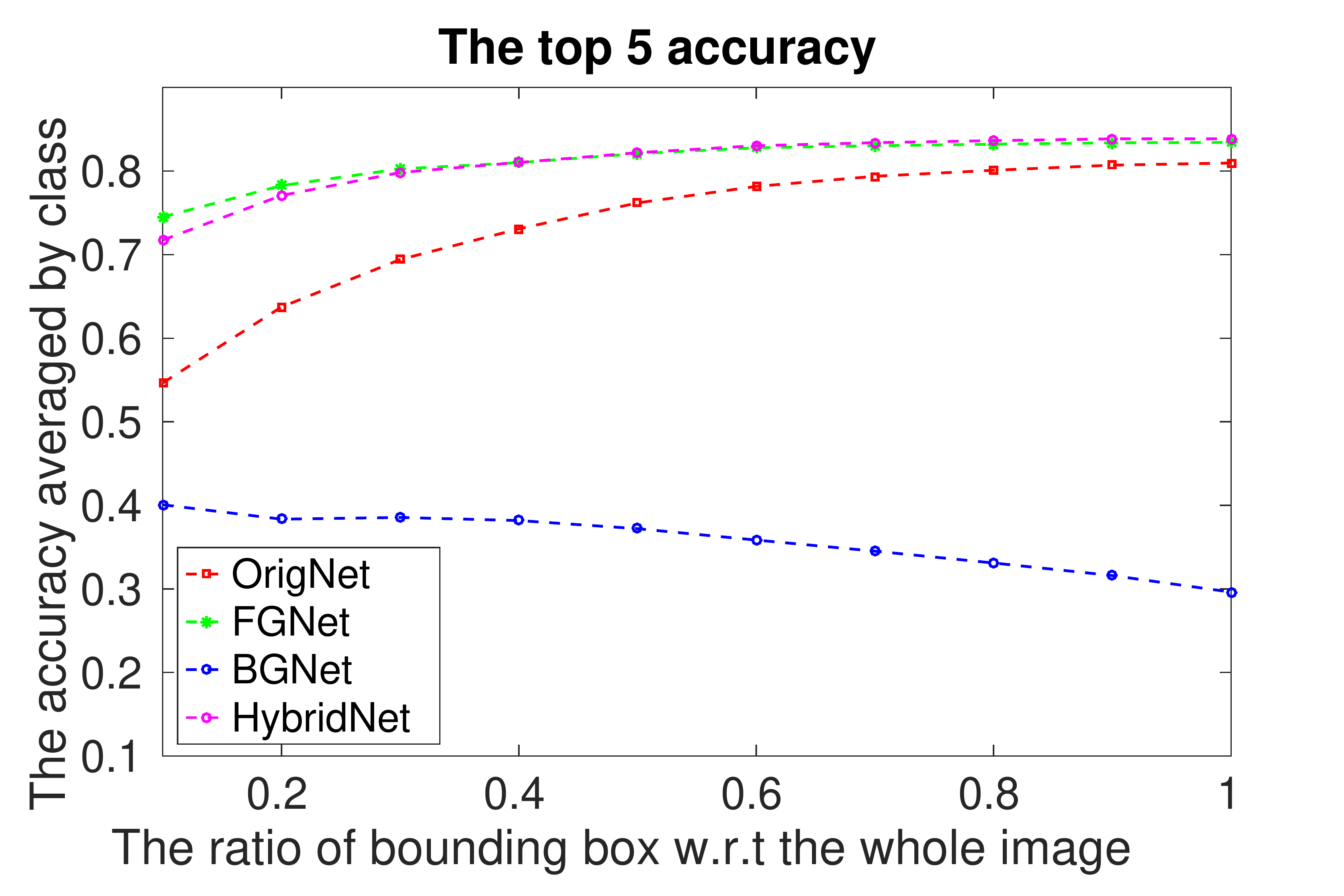}
}
\caption{
    Classification accuracy with respect to the foreground ratio on testing images.
    The number at, say, $0.3$, represents the testing accuracy on the set of all images with foreground ratio no greater than $30\%$.
    Best viewed in color.
}
\label{Fig:AccFGRatio}
\end{figure}

\subsection{Diagnosis}
\label{ComplementaryNets:Diagnosis}

We conduct diagnostic experiments to study the property of different networks to fully understand the networks behaviors. Specifically, we report the classification accuracy of different networks with respect to keeping different foreground ratios of the testing image.

We split each testing dataset into $10$ subsets, each of which contains all images with the foreground ratio no greater than a fixed value. Results are shown in Fig.~\ref{Fig:AccFGRatio}.
{\bf BGNet} gets higher classification accuracy on the images with a relatively smaller foreground ratio,
while other three networks prefer a large object ratio since the foreground information is primarily learned for recognition in these cases.
Furthermore when the foreground ratio goes larger, {\em e.g.}, greater than $80\%$,
the performance gap among {\bf OrigNet}, {\bf FGNet} and {\bf HybridNet} gets smaller.

\subsection{Visualization}
\label{ComplementaryNets:Visulization}
In this part, we visualize the networks to see how different networks learn different visual patterns.
We adopt a very straightforward visualization method~\cite{Wang_2015_Discovering}, which takes a trained network and reference images as input.

We visualize the most significant responses of the neurons on the {\em conv-5} layer.
The {\em conv-5} layer is composed of $256$ filter response maps, each of which has $13\times13$ different spatial positions.
After all the $50\rm{,}000$ reference images are processed, we obtain $13^2\times50000$ responses for each of the $256$ filters.
We pick up those neurons with the highest response and trace back to obtain its receptive field on the input image.
In this way, we can discover the visual patterns that best describe the concept this filter has learned.
For diversity, we only choose at most one patch from a reference image with the highest response score.

\begin{figure*}
\centering
\includegraphics[width=1.0\linewidth]{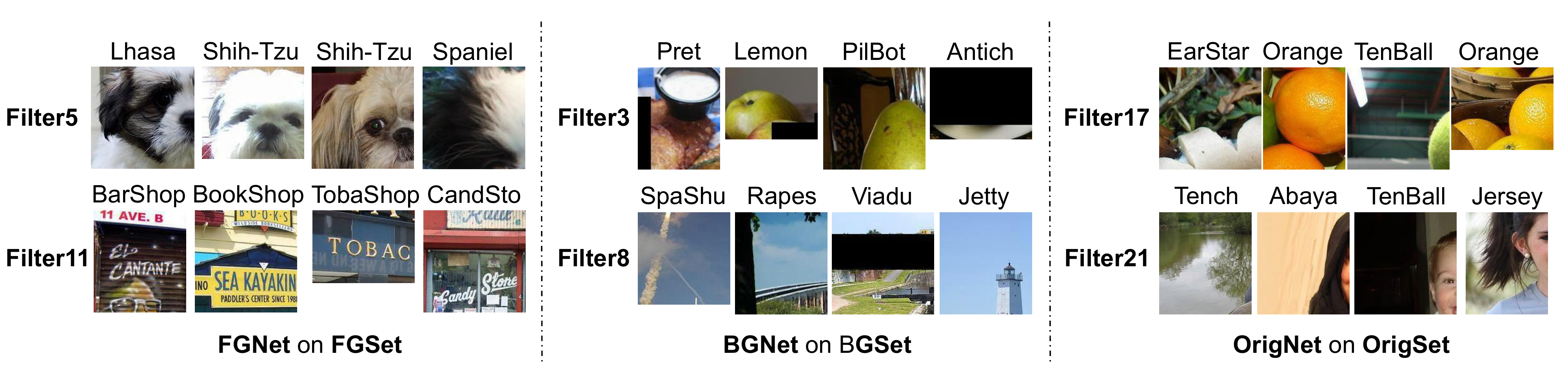}
\caption{
    Patch visualization of {\bf FGNet} on {\bf FGSet} (left), {\bf BGNet} on {\bf BGSet} (middle) and {\bf OrigNet} on {\bf OrigSet} (right).
    Each row corresponds to one filter on the {\em conv-5} layer,
    and each patch is selected from $13^2\times50000$ ones, with the highest response on that kernel.
    Best viewed in color.
}
\label{Fig:Visualization}
\end{figure*}

Fig.~\ref{Fig:Visualization} shows visualization results using {\bf FGNet} on {\bf FGSet}, {\bf BGNet} on {\bf BGSet} and {\bf OrigNet} on {\bf OrigSet}, respectively.
We can observe quite different visual patterns learned by these networks.
The visual patterns learned by {\bf FGNet} are often very specific to some object categories,
such as the patch of a {\em dog face} (filter $5$) or the {\em front side} of a {\em shop} (filter $11$).
These visual patterns correspond to some visual attributes, which are vital for recognition.
However, each visual concept learned by {\bf BGNet} tends to appear in many different object categories, for instance, the patch of {\em outdoor scene} (filter $8$) shared by the {\em jetty, viaduct, space shuttle, etc. }
These visual patterns are often found in the context, which plays an assistant role in object recognition.
As for {\bf OrigNet}, the learned patterns can be shared specific objects or scene.

To summarize, {\bf FGNet} and {\bf BGNet} learn different visual patterns that can be combined to assist visual recognition.
In Sec~\ref{ComplementaryNets:Diagnosis} we quantitatively demonstrate the effectiveness of these networks via combining these information for better recognition performance.

\section{Combination}
\label{Combination}

\begin{table}
\small
\begin{center}
\begin{tabular}{lcc}\toprule
Network                           & {\em Guided} & {\em Unguided} \\
\hline
{\bf OrigNet}                                 & $58.19\%$, $80.96\%$ & $58.19\%$, $80.96\%$ \\
{\bf BGNet}                                   & $14.41\%$, $29.62\%$ & $ 8.30\%$, $20.60\%$ \\
{\bf FGNet}                                   & $60.82\%$, $83.43\%$ & $40.71\%$, $64.12\%$ \\
{\bf HybridNet}                               & $61.29\%$, $83.85\%$ & $45.58\%$, $70.22\%$ \\
\hline
{\bf FGNet}$+${\bf BGNet}                     & $61.75\%$, $83.88\%$ & $41.83\%$, $65.32\%$ \\
{\bf HybridNet}$+${\bf BGNet}                 & $62.52\%$, $84.53\%$ & $48.08\%$, $72.69\%$ \\
{\bf HybridNet}$+${\bf OrigNet}               & $\bm{65.63}\%$, $\bm{86.69}\%$ & $\bm{60.36}\%$, $\bm{82.47}\%$ \\
\bottomrule
\end{tabular}
\end{center}
\caption{
    Classification accuracy (in terms of top-$1$, top-$5$) comparison of different network combinations.
    It's worth noting that we feed the entire image into the {\bf OrigNet} no matter whether the ground-truth bounding box(es) is given in order to keep the testing phase consistent with the training of {\bf OrigNet}. Therefore, the reported results of {\bf OrigNet} are same with each other under both {\em guided} and {\em unguided} conditions.
    To integrate the results from several networks, we weighted sum up the responses on the {\em fc-8} layer.
}
\label{Tab:LinearCombination}
\end{table}

We first show that the recognition accuracy can be significantly boosted using ground-truth bounding box(es) at the testing stage.
Next, with the help of the EdgeBox algorithm~\cite{Zitnick_2014_Edge} to generate accurate object proposals,
we improve the recognition performance without the requirement of ground-truth annotations.
We name them as {\em guided} and {\em unguided} combination, respectively.

\subsection{Guided vs. Unguided Combination}

We start with describing guided and unguided manners of the model combination.
For simplicity, we adopt the linear combination over different models, {\em i.e.}, forwarding several networks, and weighted summing up the responses on the {\em fc-8} layer.

If the ground-truth bounding box is provided (the {\em guided} condition),
we use the ground-truth bounding box to divide the testing image into foreground and background regions.
Then, we feed the foreground regions into {\bf FGNet} or {\bf HybridNet}, and background regions into {\bf BGNet}, then fuse the neuron responses at the final stage.

\begin{figure}
\centering
\includegraphics[width=1.0\linewidth]{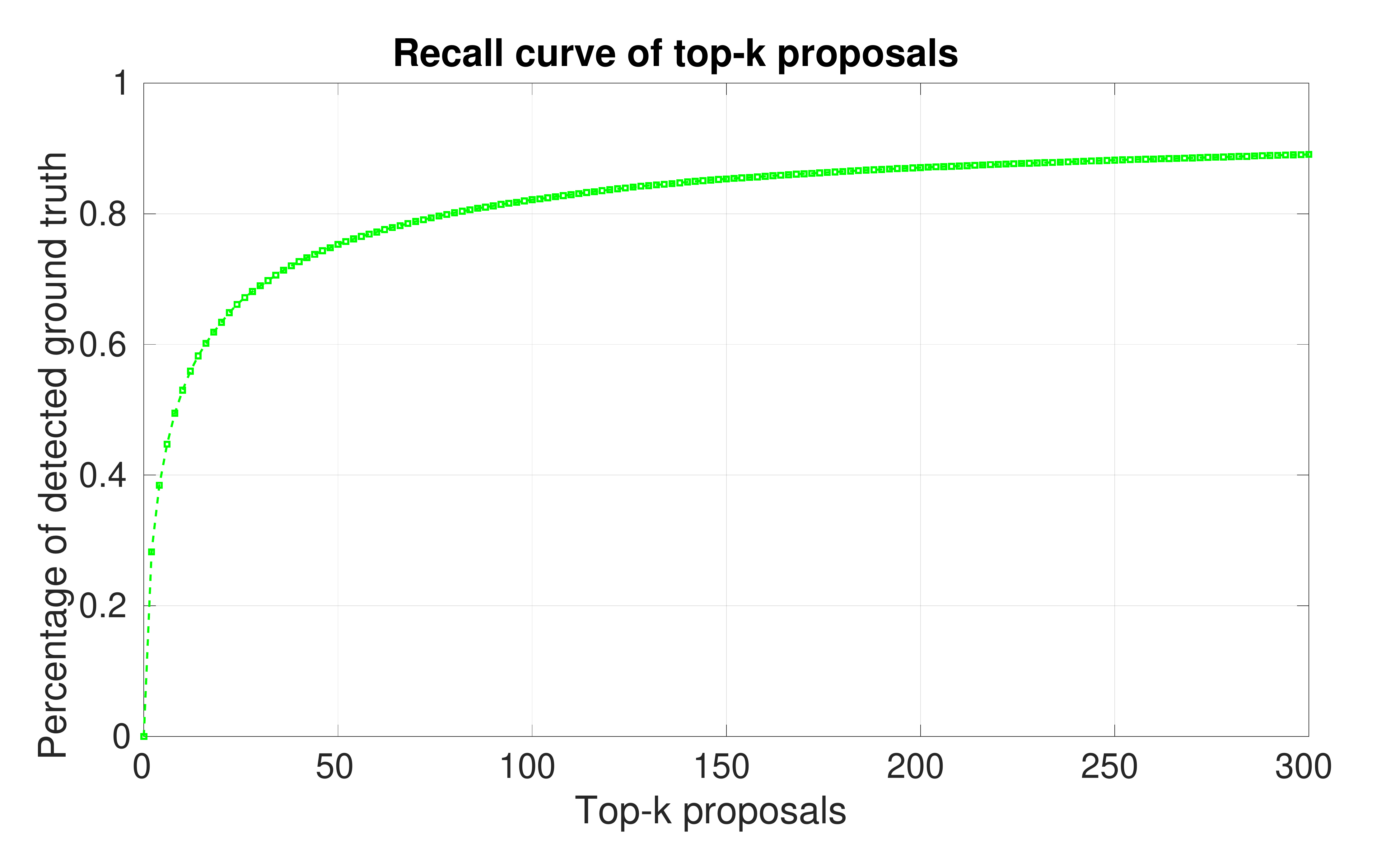}
\caption{
    EdgeBox statistics on {\bf ILSVRC2012} validation set, which denotes the curriculum distribution function of the detected ground-truth with respect to the top-$k$ proposals. Here, we set the Intersection over Union (IoU) threshold to be $0.7$ for EdgeBox algorithm.
}
\label{Fig:EdgeBoxStatistics}
\end{figure}

Furthermore, we also explore the solution of combining multiple networks in an {\em unguided} manner.
As we will see in Sec~\ref{Combination:LinearCombination}, a reliable bounding box helps a lot in object recognition.
Motivated by which, we use an efficient and effective algorithm, EdgeBox,
to generate a lot of potential bounding boxes proposals for each testing image, and then feed the foreground and background regions into neural networks as described before across top proposals.

To begin with, we demonstrate the EdgeBox proposals are good to capture the ground-truth object.
After extracting top-$k$ proposals with EdgeBox, we count the detected ground-truth if at least one of proposals has the IoU no less than $0.7$ with the ground-truth. The cumulative distribution function (CDF) is plotted in Fig.~\ref{Fig:EdgeBoxStatistics}. Considering efficiency as well as accuracy, we choose the top-$100$ proposals to feed the foreground and background into trained networks, which give an around $81\%$ recall. After obtaining $100$ outputs for each network, we average responses of {\em fc-8} layer for classification.


\subsection{Combination Results and Discussion}
\label{Combination:LinearCombination}
Results of different combinations are summarized in Table~\ref{Tab:LinearCombination}.
Under either {\em guided} or {\em unguided} settings, combining multiple networks boosts recognition performance, which verifies the statement that different visual patterns from different networks can help with each other for the object recognition.

Take a closer look at the accuracy gain under the {\em unguided} condition. The combination of {\bf HybridNet}$+${\bf BGNet} outperforms {\bf HybridNet} by $2.50\%$ and $2.47\%$ in terms of top-$1$ and top-$5$ recognition accuracy, which are noticeable gains. As for the {\bf FGNet}$+${\bf BGNet}, it improves $1.12\%$ and $1.20\%$ classification accuracy compared with the {\bf FGNet}, which are promising.
Surprisingly, the combination of {\bf HybridNet} with {\bf OrigNet} can still increase from the {\bf OrigNet} by $2.17\%$ and $1.51\%$. We hypothesize that the combination is capable of discovering the objects implicitly by the inference of where the objects are due to the visual patterns of {\bf HybridNet} are learned from images with object spatial information. One may conjecture that the performance improvement may come from the ensemble effect, which is not necessarily true considering: 1) object proposals are not accurate enough; 2) data augmentation ({\em $5$ crops} and {\em $5$ flips}) is already done for the {\bf OrigNet}, therefore the improvement is complementary to data augmentation. Moreover, we quantitatively verify that the improvements are not from simple data augmentation by giving the results of {\bf OrigNet} averaged by $100$ densely sampled patches ($50$ crops and corresponding $50$ flips, $227\times227\times3$, referred to as {\bf OrigNet100}) instead of the default ($5$ crops and $5$ flips) setting. The top-$1$ and top-$5$ accuracy of {\bf OrigNet100} are $58.08\%$ and $81.05\%$, which are very similar to original $58.19\%$ and $80.96\%$. This suggests that the effect of data augmentation by $100$ patches is negligible. By contrast, {\bf HybridNet+OrigNet100} reports $60.80\%$ and $82.59\%$, significantly higher than {\bf OrigNet100} alone, which reveals that {\bf HybridNet} brings in some benefits that are not achieved via data augmentation. These improvements are super promising considering that the networks don't know where the accurate objects are under the {\em unguided} condition. Notice that the results under {\em unguided} condition cannot surpass those under  {\em guided} condition, arguably because the top-$100$ proposals not good enough to capture the accurate ground-truth given that the {\bf BGNet} cannot give high confidence on the predictions.

For the {\em guided} way of testing, by providing accurate separation of foreground from background, works better than the {\em unguided} way by a large margin, which makes sense. And the improvements can consistently be found after combinations with the {\bf BGNet}. It is well worth noting that the combination of {\bf HybridNet} with {\bf OrigGNet} improves the baseline of {\bf OrigGNet} to a significant margin by $7.44\%$ and $5.73\%$. The huge gains are reasonable because of networks' ability to infer object locations trained on accurate bounding box(es).

\vspace{-0.2cm} 
\section{Conclusions and Future Work}
\label{Conclusions}


In this work, we first demonstrate the surprising finding that neural networks can predict object categories quite well even when the object is \textbf{\textit{not}} present. This motivates us to study the human recognition performance on foreground \textbf{\textit{with}} objects and background \textbf{\textit{without}} objects. We show on the $127$-classes {\bf ILSVRC2012} that human beings beat neural networks for foreground object recognition, while perform much worse to predict the object category only on the background without objects. Then \textbf{\textit{explicitly}} combining the visual patterns learned from different networks can help each other for the recognition task. We claim that more emphasis should be placed on the role of contexts for object detection and recognition.

In the future, we will investigate an end-to-end training approach for explicitly separating and then combining the foreground and background information, which explores the visual contents to the full extent. For instance, inspired by some joint learning strategy such as Faster R-CNN~\cite{Ren_2015_Faster}, we can design a structure which predicts the object proposals in the intermediate stage, then learns the foreground and background regions derived from the proposals separately by two sub-networks and then takes foreground and background features into further consideration.
\vspace{-0.2cm} 
\section{Acknowledgment}
This work is supported by the Intelligence Advanced Research Projects Activity (IARPA) via Department of Interior/Interior Business Center (DoI/IBC) contract number D16PC00007. We greatly thank the anonymous reviewers and JHU CCVL members who have given valuable and constructive suggestions which make this work better.
\newpage
{\footnotesize\bibliographystyle{named} 
\bibliography{egbib}}

\end{document}